\documentclass[10pt,twocolumn,letterpaper]{article}

\usepackage{arXiv}
\usepackage{times}
\usepackage{epsfig}
\usepackage{graphicx}
\usepackage{amsmath}
\usepackage{amssymb}

\usepackage{adjustbox}
\usepackage{subfig}
\usepackage{tabularx}
\usepackage{tablefootnote}
\usepackage{booktabs} %

\usepackage[numbers,sort]{natbib}
\usepackage[pagebackref=true,breaklinks=true,letterpaper=true,colorlinks,bookmarks=false]{hyperref}

\usepackage[margin=4pt,font=footnotesize,labelfont=bf,labelsep=endash,tableposition=top]{caption}

\newcommand\T{\rule{0pt}{1.2ex}}       %
\newcommand\B{\rule[-0.8ex]{0pt}{0pt}} %

\Finalcopy     
\begin{document}

	\title{Architecture Search of Dynamic Cells for Semantic Video Segmentation}
	\author{Vladimir Nekrasov
		\quad
		Hao Chen
		\quad
		Chunhua Shen
		\quad
		Ian Reid \\
		The University of Adelaide, Australia\\
		{
			\small
			E-mail:
			\{vladimir.nekrasov, hao.chen01, chunhua.shen, ian.reid\}@adelaide.edu.au 
		}
	}
	
	\maketitle

	\begin{abstract}
		In semantic video segmentation the goal is to acquire consistent dense semantic labelling across image frames. To this end, recent approaches have been reliant on manually arranged operations applied on top of static semantic segmentation networks - with the most prominent building block being the optical flow able to provide information about scene dynamics. Related to that is the line of research concerned with speeding up static networks by approximating expensive parts of them with cheaper alternatives, while propagating information from previous frames. In this work we attempt to come up with generalisation of those methods, and instead of manually designing contextual blocks that connect per-frame outputs, we propose a neural architecture search solution, where the choice of operations together with their sequential arrangement are being predicted by a separate neural network. We showcase that such generalisation leads to stable and accurate results across common benchmarks, such as CityScapes and CamVid datasets. Importantly, the proposed methodology takes only $2$~GPU-days, finds high-performing cells and does not rely on the expensive optical flow computation.
	\end{abstract}

	\section{Introduction}
	
	Human beings are well-equipped by evolution to quickly observe changes in dynamic environments. From merely few seconds of studying an unknown scene, we are able to coherently map out its main constituents. In contrast, static semantic segmentation networks would perform poorly in such conditions, and may as well produce contradictory predictions across the frames. Therefore, the question arises of how to make the static models suitable for segmenting continuously evolving scenes? 
	
	One well-known approach would be to use the optical flow that describes the motion in the scene between adjacent frames~\cite{GaddeJG17,ZhuXDYW17}. The optical flow calculation tends to be expensive and also comes with several notable disadvantages, among which its inability to deal with occlusions and newly appeared objects. Nevertheless, as shown by Gadde~\etal~\cite{GaddeJG17}, a relatively poor estimate of the optical flow may still carry significant benefits, not the least of which lies in computational savings. 
	
	Alternatively, one may choose to model which information must be propagated across the frames, \eg with the help of a recurrent neural network with memory units~\cite{NilssonS18}. Even more biologically plausible are the models that compute different features at various time-scales~\cite{ShelhamerRHD16}, in a vein similar to neural spikes. Naturally, this comes with its own set of disadvantages, most notably the difficulty of choosing an appropriate scheduling regime for updating individual parts of the network.
	
	Yet another complementary line of work focuses on approximating an expensive per-frame forward pass with cheaper alternatives: \eg Li~\etal~\cite{LiSL18} predicted local filters to be applied on the segmentation prediction from the previous frame, while Jain~\etal~\cite{abs-1807-06667} used a larger network for key frames and directly employed a smaller one for consecutive frames. Such savings may allow to re-use more expensive optical flow methods without a significant slowdown, but the choice of key frames can be crucial and not readily justifiable.
	
	\begin{figure}[t]
		\begin{center}
			\includegraphics[width=0.98\linewidth]{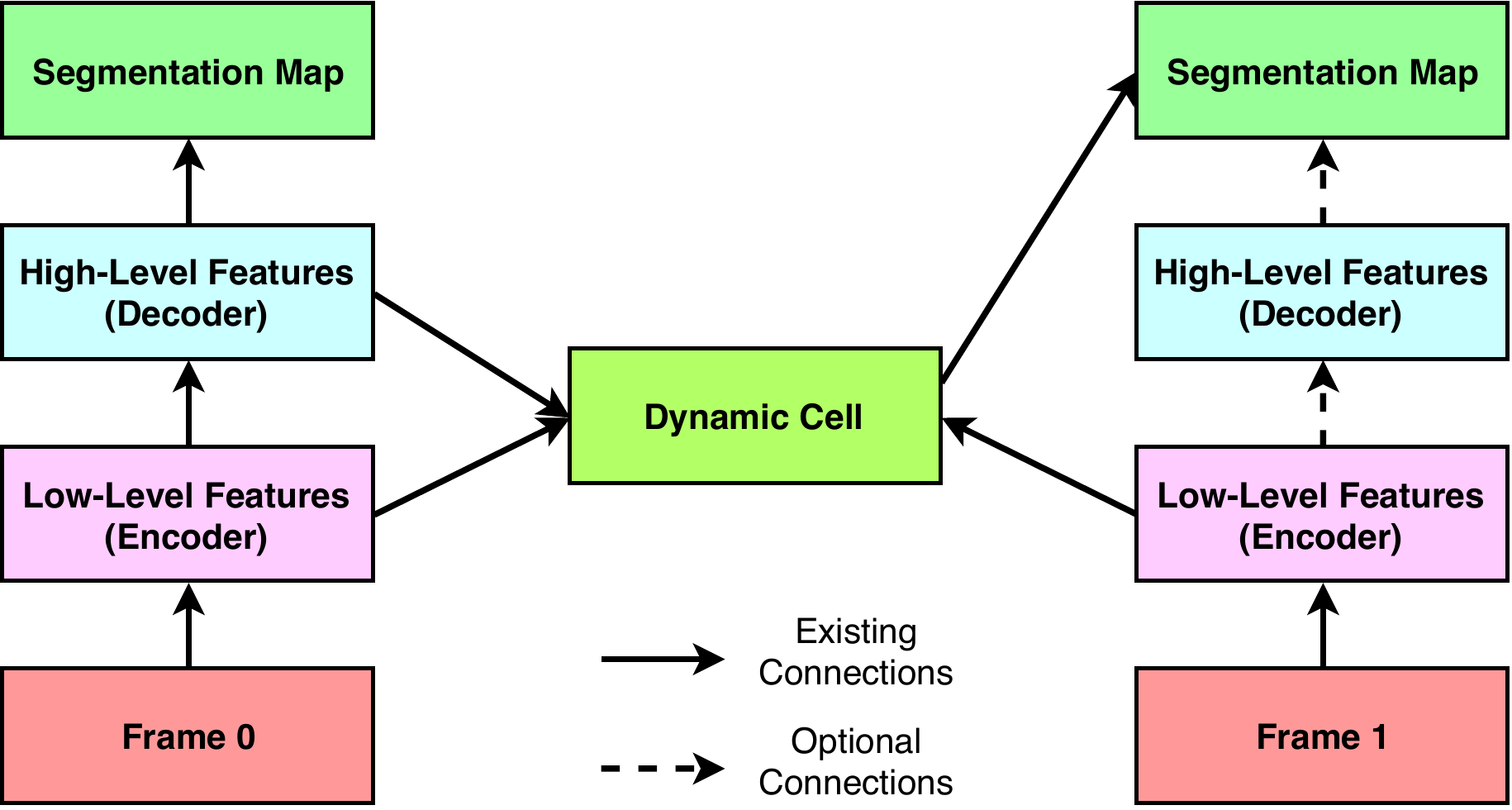}
		\end{center}
		\caption{Semantic video segmentation approaches tend to comprise a dynamic cell that takes as inputs the information from the previous and current frames, and outputs the segmentation mask. For example, the dynamic cell can calculate the optical flow~\cite{GaddeJG17}, or predict convolutional filters~\cite{LiSL18}. In this work we use NAS to discover novel and high-performing dynamic cells.\label{fig1}}
	\end{figure}
	
	Looking closely at the aforementioned approaches for video semantic segmentation, one may notice an easily discernible pattern: a typical video segmentation network predicts a labelling of the current frame based on the information propagated from the previous one and hidden representations of the current one~(Fig.~\ref{fig1}). While seemingly obvious, it possesses certain variations depending on the goal - \eg whether efficiency, or real-time performance is desired. Importantly, what we would like to emphasise here is that, while technically sound, all the current approaches have been manually designed and have not considered any interplay between different building blocks.
	
	Starting from that general pattern we instead propose to leverage the neural architecture search ({\it NAS})~\cite{ZophL16} methodology to find contextual blocks that enhance a per-frame segmentation network with dynamic components. This motivation is justified by recent results achieved using NAS on such tasks as image classification~\cite{ZophVSL17,LiuZNSHLFYHM18}, language modelling~\cite{PhamGZLD18} and static semantic segmentation~\cite{abs-1809-04184,abs-1810-10804}, that oftentimes outperform manually designed networks. We build upon those results and adapt current approaches in a way suitable for handling the dynamic nature of dense per-pixel classification. To the best of our knowledge, we are the first to consider the application of NAS to the task of video semantic segmentation.
	
	Our automated approach comes with certain benefits, concretely:
	\begin{itemize}
		\itemsep -.122cm
		\item[i.)] 
		it considers a larger span of initial building blocks than any previous work, 
		\item[ii.)] 
		it empirically evaluates different design structures and finds most promising ones, and
		\item[iii.)]
		it requires only few GPU-days to find a set of high-performing structures.
	\end{itemize}
	Furthermore, although we do not consider it in this work, the proposed methodology can further be extended to take into account different specific objectives (even non-differentiable), such as runtime~\cite{abs-1812-09926}.
	
	\section{Related Work}
	
	\subsection{Static semantic segmentation} Most recent approaches in static semantic segmentation have been exploiting fully convolutional neural networks~\cite{LongSD15}. Typical methods are based either on the encoder-decoder structure with skip-connections~\cite{LongSD15,LinMSR17}, dilated convolutional layers~\cite{YuKF17,ZhaoSQWJ17,ChenPKMY18}, or the combination of the above~\cite{ChenZPSA18}. Per-frame instantiations of these networks are usually computationally expensive, hence, several works have considered building light-weight segmentation architectures~\cite{ZhaoQSSJ18,NekrasovS018}. Nevertheless, due to the lack of information propagation between frames, these networks perform poorly on videos and are unable to provide consistent results.
	
	\subsection{Dynamic semantic segmentation} One of the first lines of work in video segmentation has been built upon the usage of the optical flow~\cite{ZhuXDYW17}, in which features extracted from the previous frame are propagated to the current one via warping. This usually results in a slight computational overhead, although as noted by Gadde~\etal~\cite{GaddeJG17} an easily attainable noisy estimate of the optical flow still carries significant benefits. Nevertheless, the optical flow does not fair well in situations when scenes are undergoing substantial changes with novel objects constantly appearing and multiple occlusions being present. Thus, Jain~\etal~\cite{abs-1807-06667} have proposed to combine the optical flow estimate with a relatively cheaper approximation of the current frame using a smaller network. Xu~\etal~\cite{XuFYL18} have chosen to assign different image regions to two different networks to process: while the first one - deep and slow - works on regions that have significantly changed, the second one - shallow - predicts new features based on the optical flow information. In a similar vein, Nilsson and Sminchisescu~\cite{NilssonS18} have propagated labels from the previous frame at only those pixels where the optical flow estimate is reliable. 
	
	A seemingly different approach, proposed by Li~\etal~\cite{LiSL18}, instead predicts local convolutional kernels based on the low-level representation of the current frame that are applied on the prediction from the previous frame. Importantly, while the current estimate is being used for next frame, a more accurate one is being computed in parallel for future re-use.
	
	In yet another line of work, Chandra~\etal~\cite{ChandraCK18} have adapted Deep Gaussian Random Field~\cite{ChandraK16} to handle temporal information by predicting besides unary and spatial pairwise terms also temporal pairwise terms, efficiently propagating features between frames.

	\subsection{Neural Architecture Search} NAS methods aim to find high-performing architectures in an automated way. Here, we consider the reinforcement learning-based (RL) approach~\cite{ZophL16}, where a separate recurrent neural network (controller) outputs a sequence of tokens describing an architecture that should provide highest score on the holdout validation set.
	
	While there is no prior work on NAS for video segmentation, two results in static segmentation are worth mentioning: Chen~\etal~\cite{abs-1809-04184} used a random search to find a single set of operations (so-called `cell') on the top of the DeepLab architecture~\cite{ChenPKMY18}, while Nekrasov~\etal~\cite{abs-1810-10804} exploited RL to find a cell together with the topological structure of the encoder-decoder type of architecture. We borrow one of the architectures found by Nekrasov~\etal as our static baseline, and extend their NAS approach for video segmentation. Since we are only searching for the dynamic component that connects different instantiations of the already pre-trained static segmentation network, we are able to train and evaluate each candidate in a short amount of time, the trait that is extremely important for all NAS methods.

	\section{Methodology}
	
	As noted in introduction and depicted in Fig.~\ref{fig1}, we attempt to generalise previous solutions for video semantic segmentation in such a way that NAS methods become readily applicable. To this end, we look for a single cell that connects representations from the previous frame and enhances current predictions without a significant overhead. What follows is the description of the input space (Sect.~\ref{inp-space}), the search space (Sect.~\ref{search-space}), and the search approach (Sect.~\ref{search}).
	
	\begin{figure}[t]
		\begin{center}
			\includegraphics[width=0.95\linewidth]{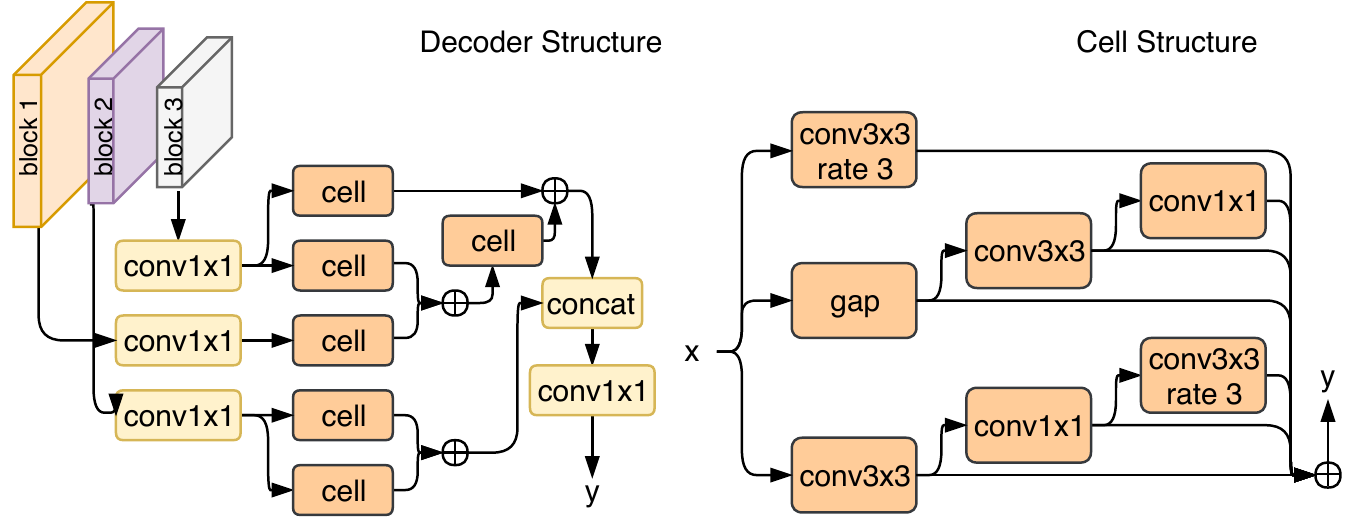}
		\end{center}
		\caption{\emph{arch2}~\cite{abs-1810-10804}. \emph{`gap'} stands for global average pooling.\label{arch2}}
	\end{figure}
	
	\subsection{Input space}
	\label{inp-space}
	We consider the {\em arch2} network from the work of Nekrasov~\etal~\cite{abs-1810-10804}. It is an encoder-decoder type of the segmentation network with the encoder being a light-weight classifier (MobileNet-v2~\cite{abs-1801-04381}), and the decoder being an automatically discovered structure presented in Fig.~\ref{arch2}. This architecture strikes a fine balance between accuracy and runtime, both being important characteristics for semantic video segmentation. Here it should be noted that the application of our methodology is not directly tied to a concrete architecture and can be easily adapted to work with other networks.
	
	In the proposed setup, the static network is applied end-to-end on the first frame and three outputs are being recorded: an intermediate representation - in this case, the encoder's output with the resolution of $\frac{1}{32}$ of the input image ($layer~4$), the decoder's output before ($dec$) and after the final classifier ($pred$) - both with resolutions of $\frac{1}{8}$ of the input image and with $64$ and $C$ numbers of channels, correspondingly, where $C$ is the number of output classes. For the second frame, we record three outputs from the encoder only - two intermediate ones with the resolutions of $\frac{1}{8}$ ($layer~2$) and $\frac{1}{16}$ ($layer~ 3$), respectively, and the final one with the resolution of $\frac{1}{32}$ ($layer~4$).
	
	We rely on the dynamic cell, the layout of which will be described below, to predict the semantic labelling of the current frame given $5$ inputs: $layer~4$ and $dec$ from the previous frame, and $layers~2{-}3{-}4$ from the current one. This way, we do not have to execute the decoder part of the static segmentation network on the current frame (thus decreasing latency), at the same time re-using information from the previous frame. The output of the dynamic cell serves as the input $dec$ for the next frame.

	\subsection{Search space}
	\label{search-space}
	We rely on an LSTM-based controller to predict a sequence of operations together with locations where they should be applied in order to form a dynamic cell~\cite{abs-1810-10804}. Concretely, we first choose two layers out of the provided five (with replacement), two corresponding operations that need to be applied on each of them, and an aggregation operation that combines two inputs into a single output. On the next step, we repeat this process, but now we are sampling two layers out of six possible, with the aggregated result being added into the sampling pool. This process can be repeated multiple times, with the final output being formed by the concatenation of all non-sampled aggregated results.
	
	We rely on a similar set of operations as for static segmentation (Table~\ref{table:ops}), and in order to enable the dynamic cell to apply convolutional filters on irregular grids, we also include deformable $3\times3$ convolution~\cite{abs-1811-11168}.
	
	\begin{table}[htb]
		\begin{center}
			\begin{tabularx}{0.5\textwidth}{c|X}
				\specialrule{.15em}{0em}{0em}
				ID & Description \T\B\\
				\specialrule{.05em}{0em}{0em}
				\hline
				0 & separable conv $3\times3$\T\B\\
				\hline
				1 & global average pooling followed by upsampling and conv $1\times1$\T\B\\
				\hline
				2 & separable conv $3\times3$ with dilation rate $3$\T\B\\
				\hline
				3 & separable conv $5\times5$ with dilation rate $6$\T\B\\
				\hline
				4 & skip-connection\T\B\\
				\hline
				5 & deformable $3\times3$ convolution\B\\
				\specialrule{.15em}{0em}{0em}
			\end{tabularx}
			\caption{Description of operations used in the search process.
				\label{table:ops}}
		\end{center}
	\end{table}
	
	While Nekrasov~\etal~\cite{abs-1810-10804} simply summed up two different inputs at each step, here to compensate for the dynamic nature of our problem we consider a set of aggregation operations given in Table~\ref{table:agg-ops}.

	\begin{table}[htb]
	\begin{center}
		\begin{tabularx}{0.5\textwidth}{c|X}
			\specialrule{.15em}{0em}{0em}
			ID & Description \T\B\\
			\specialrule{.05em}{0em}{0em}
			\hline
			0 & summation with per-channel learnable weights per each input\T\B\\
			\hline
			1 & channel-wise concatenation of two inputs followed by conv $1\times1$ to reduce the number of channels to the original size\T\B\\
			\hline
			2 & (weight) predictive operation, where the first input becomes a set of spatial convolutional filters (weights) applied on the second one\T\B\\
			\hline
			3 & bilinear sampling of the first input, where an affine grid is predicted based on the values of the second input~\cite{JaderbergSZK15}\T\B\\
			\hline
			4 & 3D-convolution where two inputs are stacked together forming a new dimension with $2\times3\times3$ convolution applied on top\T\B\\
			\hline
			5 & dense attention: \ie element-wise multiplication between the first input and the sigmoid-activated second one\B\\
			\specialrule{.15em}{0em}{0em}
		\end{tabularx}
		\caption{Description of aggregation operations used in the search process.
			\label{table:agg-ops}}
	\end{center}
	\vspace{-8mm}
	\end{table}

	Based on the previous works, we conjecture that this set of operations will be sufficient for the task of video segmentation, and we provide experimental results to support this claim.
	
	\subsection{Finding optimal architectures}
	\label{search}
	We assume that there exists a video dataset that comes with segmentation annotations for at least a subset of consecutive frames. From it, we build pairs (or triplets) of frames such that in each sequence all the frames following the first one are always annotated. As commonly done, we further divide this set into two disjoint parts - meta-train and meta-val. We further assume an existence of the static segmentation network pre-trained on this dataset\footnote{Please refer to Sect.~\ref{static:train} for the details on pre-training of static segmentation networks.} - in particular, {\em arch2} from~\cite{abs-1810-10804}. As mentioned above, we chose this particular architecture due to its compactness and low latency.
	
	The controller samples a structure of the dynamic cell which we train on the meta-train set and evaluate on meta-val. As done in \cite{abs-1810-10804}, we consider the geometric mean of three metrics as the validation score: mean intersection-over-union ({\it mIoU}), frequency-weighted IoU ({\it fwIoU}) and mean-pixel accuracy ({\it mAcc}). This score is used by the controller to update its weights, and the process is repeated multiple times. After that, one can either sample several cells from the trained controller, or simply choose best found cells that achieved highest results during the search process.

	\section{Experiments}
	
	We conduct all our experiments on two popular video segmentation benchmark datasets - CamVid~\cite{BrostowFC09} and CityScapes~\cite{CordtsORREBFRS16}.
	
	The first one, {\em CamVid}, comprises $701$ RGB images of resolution $480{\times}360$ densely annotated into $11$ categories. Following previous work~\cite{BadrinarayananK17}, we use the dataset splits of $367$ images for training, $101$ - for validation and $233$ - for testing. We train generated architectures with batches of examples each comprising $3$ consecutive frames. 
	
	The {\em CityScapes} dataset contains $5000$ high-resolution $2048{\times}1024$ images densely labelled with $19$ semantic classes - $2975$ for training, $500$ for validation and $1525$ for testing, respectively. In addition to that, raw unannotated frames extracted from videos are also provided. For each annotated example, we add an image frame that precedes it and train architectures with batches of sequences of length $2$, in which the second frame is always annotated.
	
	In each case, we initialise the decoder's output {\em dec} on the first frame using the pre-trained static segmentation network, and rely on the dynamic cell at all following frames in the sequence as described in Sect.~\ref{inp-space}. To update the dynamic cell weights, we sum up cross-entropy loss terms at each frame after the first one and back-propagate the gradients.
	
	For both, search and training, we exploit a single V$100$ GPU with $32$GB of memory.
	
	\subsection{Search}
	
	For searching we only employ the training splits of each dataset. We further divide each randomly in $2$ non-overlapping sets - meta-train ($90\%$) and meta-val ($10\%$). We pre-compute all required outputs from the pre-trained static network and store them in memory. The static network is kept unchanged during the whole search process. Each generated architecture is trained on the meta-train split and evaluated on meta-val. We keep track of average performance and apply early stopping halfway through the training if the generated architecture is un-promising as done in~\cite{abs-1810-10804}.
	
	Our controller is a two-layer LSTM with $100$ hidden units randomly initialised from uniform distribution~\cite{abs-1810-10804}. The controller is trained with PPO~\cite{SchulmanWDRK17} with the learning rate of $1{e-}4$. To reduce the size of generated cells, we set the number of emitted layers (each layer is a string of five tokens as described in Sect.~\ref{search-space}) to $5$ on CamVid and to $4$ on CityScapes.
	
	For {\em CamVid}, we train predicted cells on mini-batches of $48$ sequences for $20$ epochs with the learning rate of $8{e-}3$ and the Adam learning rule~\cite{KingmaB14}. Each image--segmentation mask pair in the sequence is cropped to $600$ with the shorter side being mean-padded to $400$. No transformations are applied to the validation sequences.
	
	For {\em CityScapes}, we train for $10$ epochs with $48$ sequences each cropped to $512{\times}512$ with the longer side being resized to $1024$. 
	
	\subsubsection*{Results}

\begin{figure}[htb]
		\begin{center}
			\includegraphics[width=1.0\linewidth]{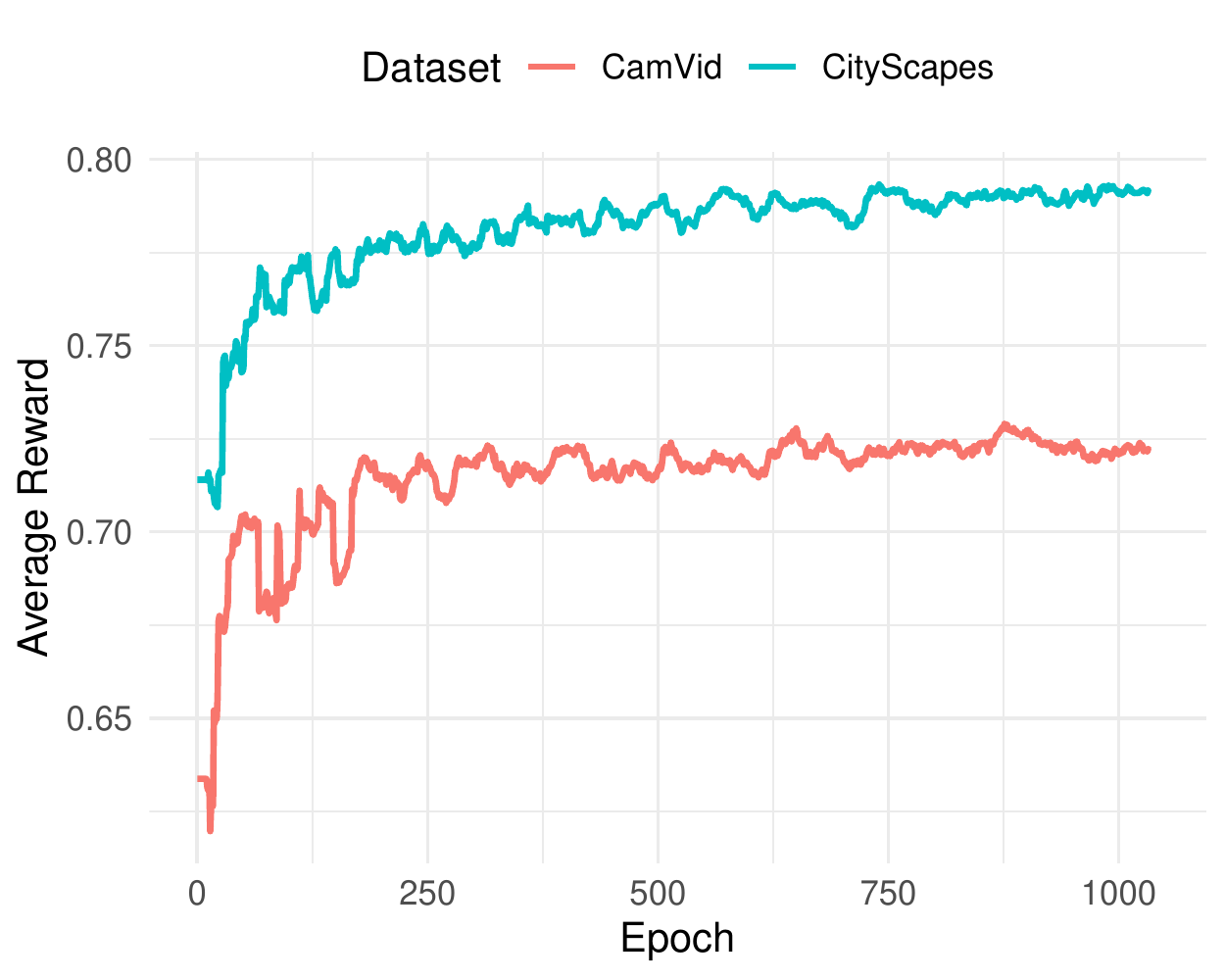}
		\end{center}
		\caption{Average search rewards on CamVid and CityScapes datasets.\label{fig:rewards}}
	\end{figure}

	We visualise the progress of rewards on each dataset in Fig.~\ref{fig:rewards}. Although the rewards are not directly comparable between the datasets, the growth dynamics on both datasets signal that the controller is able to discover better architectures throughout the search process.
	
\begin{figure*}[t]
		\subfloat[Operations\label{fig:prop-op}]{%
			\begin{minipage}{0.33\linewidth}
				\centering
				\includegraphics[width = 1.\linewidth]{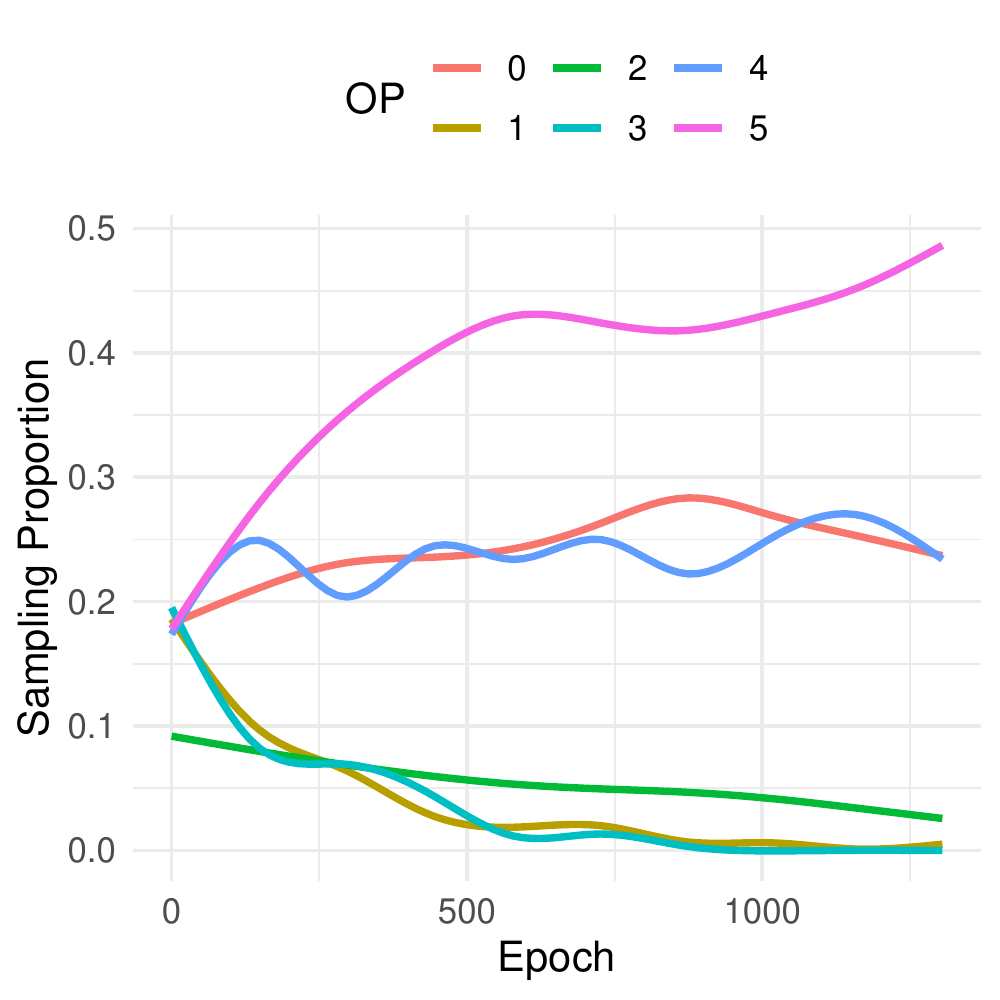}%
			\end{minipage}%
		}
		\subfloat[Aggregation Operations\label{fig:prop-agg-op}]{%
			\begin{minipage}{0.33\linewidth}
				\centering
				\includegraphics[width = 1.\linewidth]{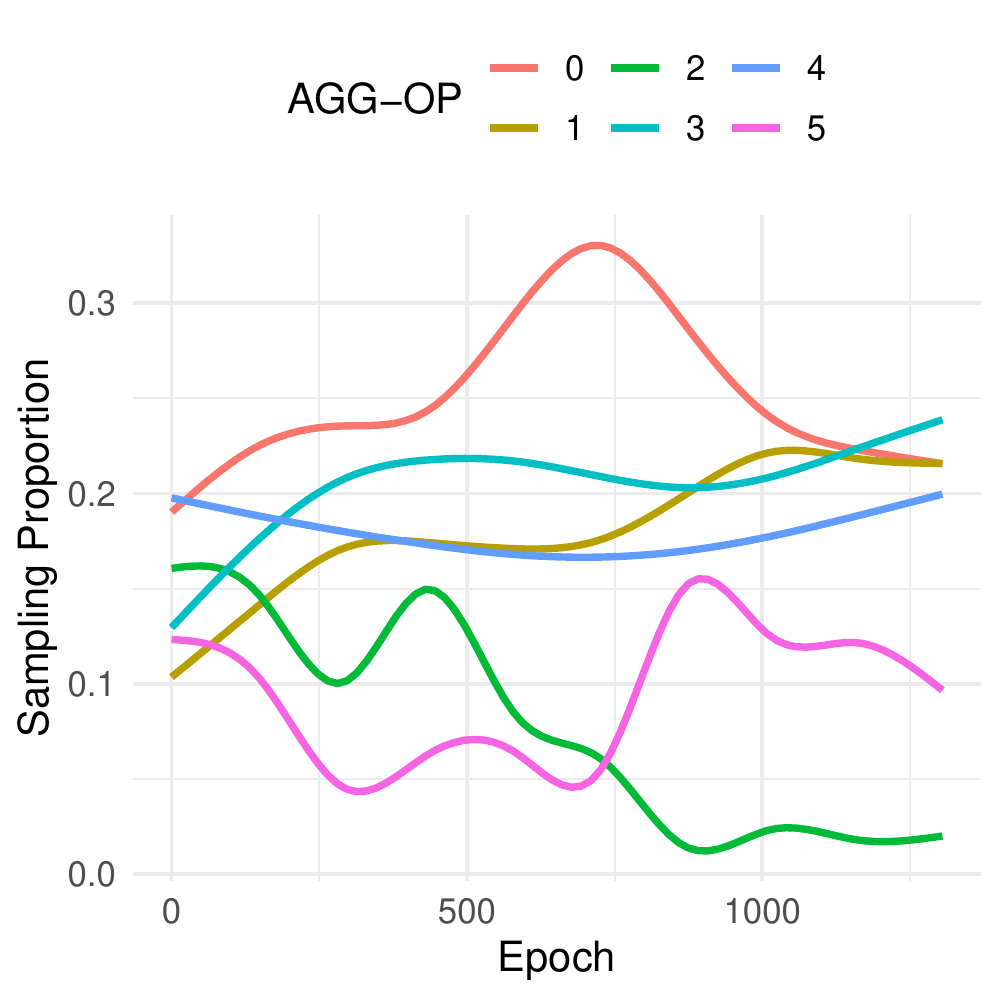}%
			\end{minipage}%
		}
		\subfloat[Input layers\label{fig:prop-inputs}]{%
			\begin{minipage}{0.33\linewidth}
				\centering
				\includegraphics[width = 1.\linewidth]{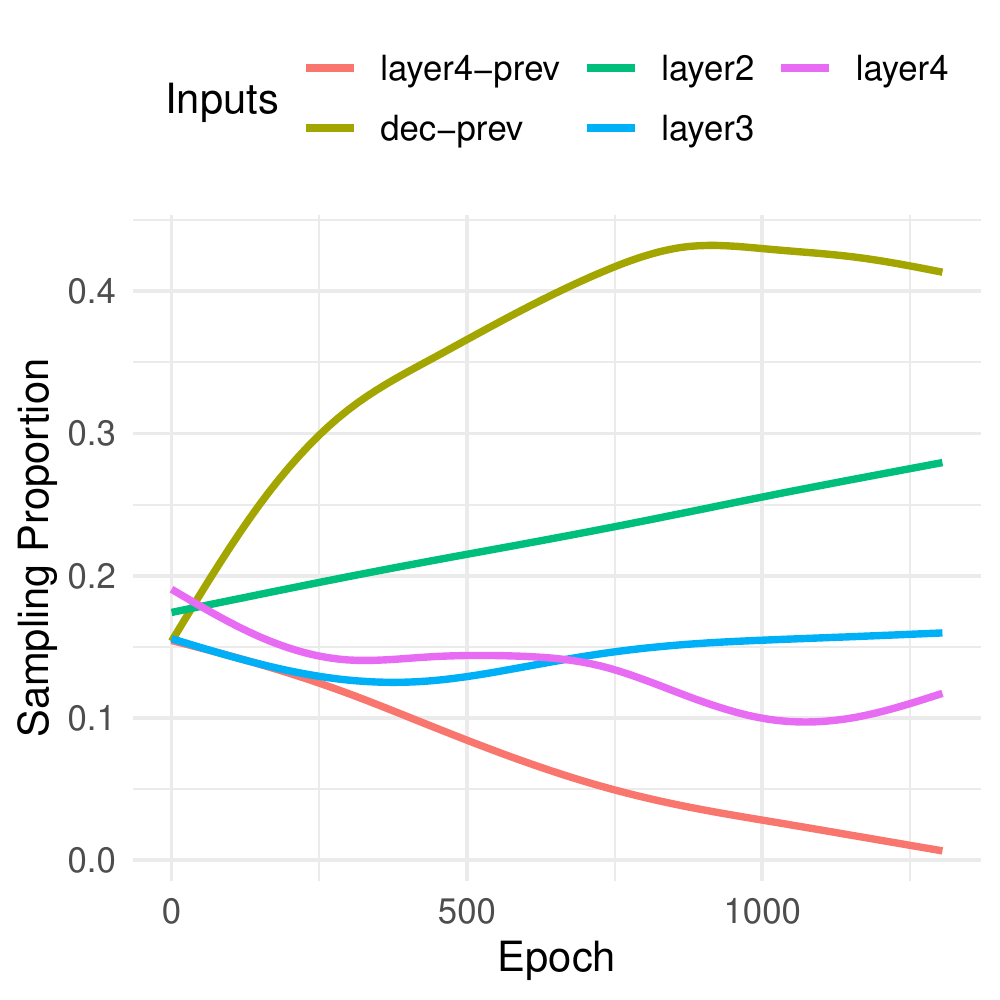}%
			\end{minipage}%
		}
		\hfill
		\subfloat[Operations\label{fig:cv-prop-op}]{%
			\begin{minipage}{0.33\linewidth}
				\centering
				\includegraphics[width = 1.\linewidth]{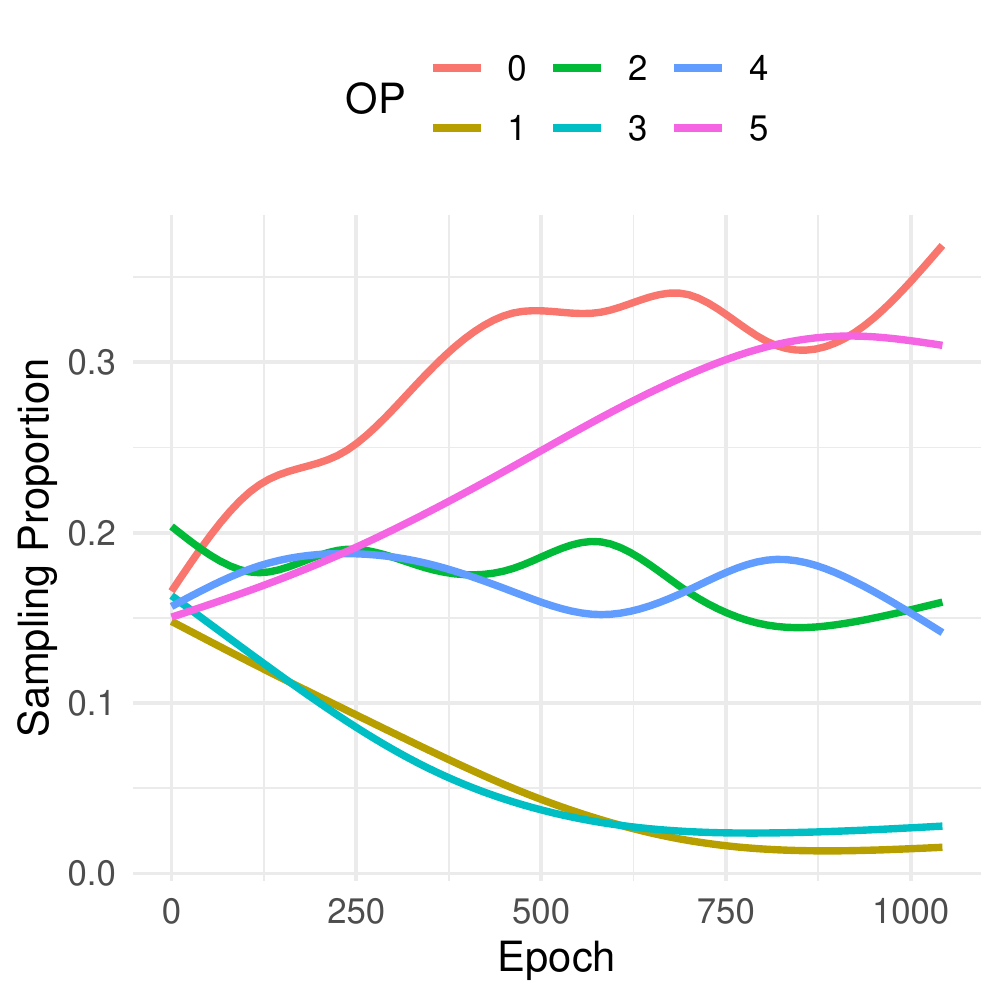}%
			\end{minipage}%
		}
		\subfloat[Aggregation Operations\label{fig:cv-prop-agg-op}]{%
			\begin{minipage}{0.33\linewidth}
				\centering
				\includegraphics[width = 1.\linewidth]{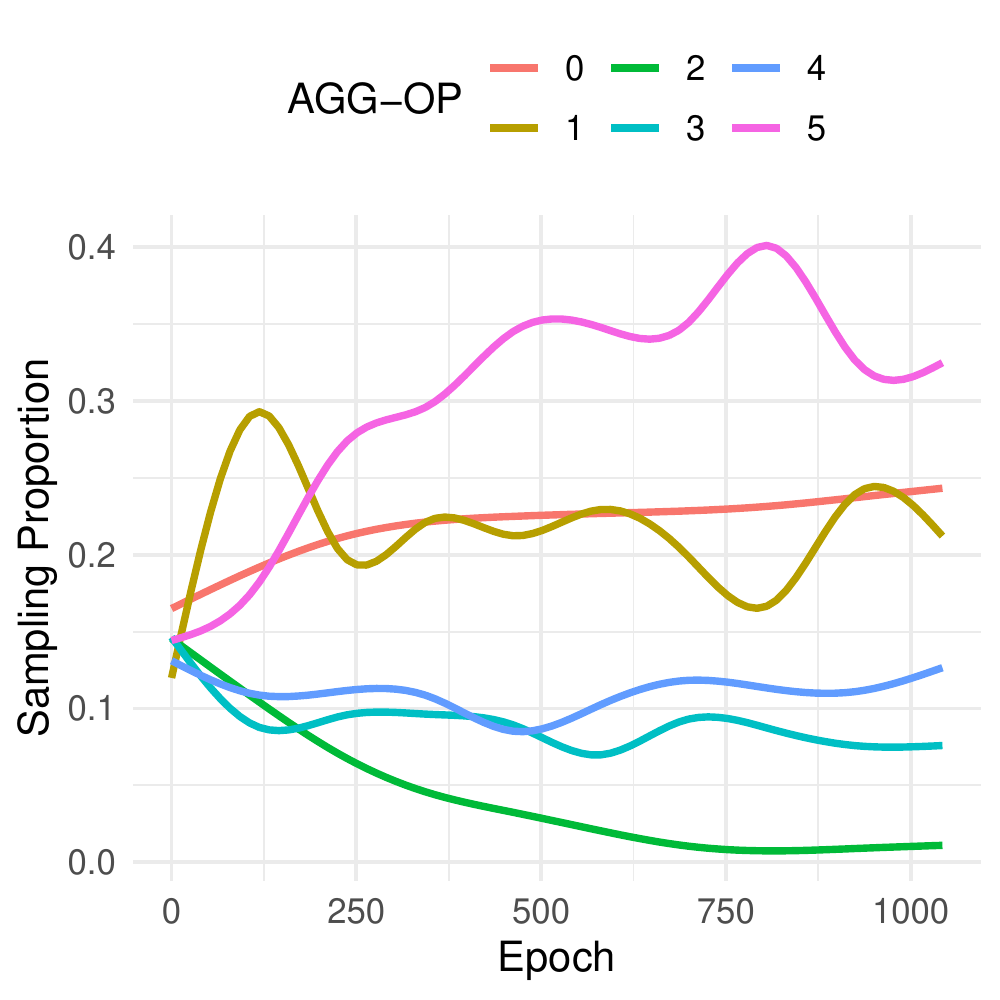}%
			\end{minipage}%
		}
		\subfloat[Input layers\label{fig:cv-prop-inputs}]{%
			\begin{minipage}{0.33\linewidth}
				\centering
				\includegraphics[width = 1.\linewidth]{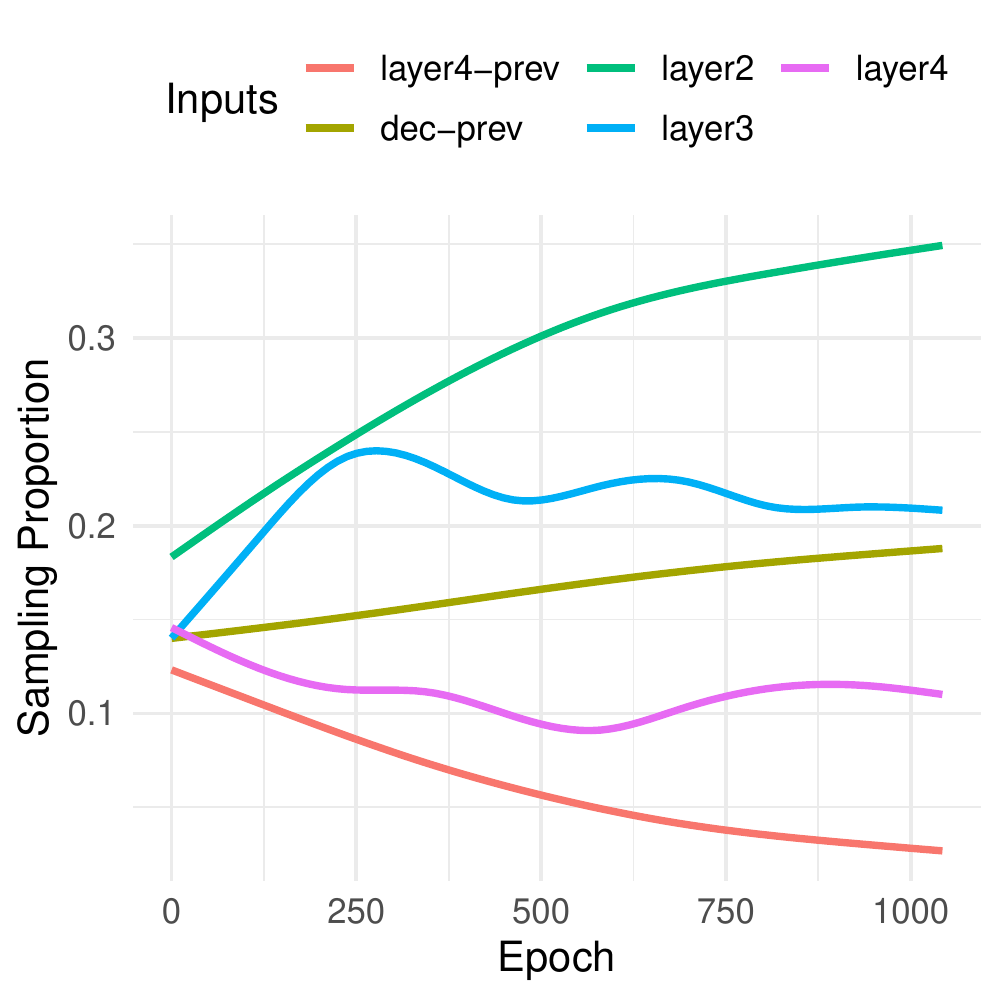}%
			\end{minipage}%
		}
		\caption{Average sampling proportion of operations, aggregation operations and input layers on CityScapes (\textbf{a-c}) and CamVid (\textbf{d-f}). Please refer to Tables~\ref{table:ops} and~\ref{table:agg-ops} for the description of operations.
			\label{fig:search-prop}}
	\end{figure*}

	We further look at the distributions of sampled operations, aggregation operations and input layers plotted on Fig.~\ref{fig:search-prop}. On both datasets, global average pooling and separable $5{\times}5$ convolution with dilation rate $6$ are sampled less frequently than other operations, potentially indicating that these layers could be omitted from the search process. On average, the controller trained on CityScapes prefers sampling deformable convolution (Fig.~\ref{fig:prop-op}), while the CamVid one - separable $3{\times}3$ convolution (Fig.~\ref{fig:cv-prop-op}).
	
	In terms of aggregation operations, the dynamics between two controllers vary significantly: the CamVid-based controller tend to rely on dense attention, while omitting the predictive operation~(Fig.~\ref{fig:cv-prop-agg-op}). In contrast, the CityScapes controller is more likely to apply bilinear sampling on an affine grid, and to ignore predictive operation together with dense attention~(Fig.~\ref{fig:prop-agg-op}).
	
	When sampling the input layers, the controllers behave similarly: in particular, both tend to skip $layer~{4}$ from the previous and current frames. The CityScapes controller extensively uses information from the previous $dec$ layer~(Fig.~\ref{fig:prop-inputs}), while the CamVid one - from $layer~{2}$ of the current frame~(Fig.~\ref{fig:cv-prop-inputs}). This may well imply that on CityScapes the final predictions on the current frame change only slightly with respect to the previous frame.
	
	Importantly, these observations indicate that two controllers trained on two different datasets exhibit various patterns, potentially capturing dataset-specific attributes in order to discover better performing architectures.

	\subsection{End-to-end Training}
	
	We further select top-$2$ performing dynamic cells on each dataset to train end-to-end on full training sets for longer.
	
	In particular, for {\em CamVid}, we pre-train the dynamic cell only with Adam and the learning rate of $8{e-}3$ for $10$ epochs with the batch size of $16$ sequences. Then we decrease the cell's learning rate in half, and fine-tune the whole architecture (i.e., with the per-frame segmentation network) end-to-end for $100$ epochs - the static network weights are updated using SGD with momentum of $0.9$ and the learning rate of $5{e-}4$. Each sample in the batch is cropped to $600{\times}600$ with the shorter side being padded to $400$.
	
	On {\em CityScapes} we pre-train for $200$ epochs with the batch size of $16$ sequences and fine-tune end-to-end for $200$ epochs. Each example in the batch is cropped to $769{\times}769$.
	
	\subsubsection*{CamVid Results}
	
	We provide quantitative results on CamVid in Table~\ref{table:camvid}. The inclusion of dynamic cells in both cases leads to an improvement over baseline by more than $1\%$. Importantly, with the exclusion of first frame in the sequence, we do not rely on expensive computations involving the static decoder.
	
	Both our models perform comparably to other state-of-the-art video segmentation networks even though the backbone that we rely on - MobileNet-v2~\cite{abs-1801-04381} - is much smaller in comparison to ResNet-101~\cite{HeZRS16} exploited by Chandra~\etal~\cite{ChandraCK18}, or DilatedNet~\cite{corr/YuK15} - by Gadde~\etal~\cite{GaddeJG17} and GRFP~\cite{NilssonS18}. Furthermore, we did not make any use of higher-resolution images of $960{\times}720$ to further improve our scores.
	
	\begin{table}[htb]
		\begin{center}
			\begin{adjustbox}{max width=0.45\textwidth}
				\begin{tabular}{l|c}
					\specialrule{.15em}{0em}{0em}
					Method & mIoU,\% \T\B\\
					\specialrule{.1em}{0em}{0em}
					\hline
					per-frame baseline & 65.3 \T\B\\
					w/ cell0 & 66.6\T\B\\
					w/ cell1 & 66.9\T\B\\
					\hline
					GRFP~\cite{NilssonS18} & 66.1\T\B\\
					Chandra~\etal~\cite{ChandraCK18} & 67.0\T\B\\
					Gadde~\etal~\cite{GaddeJG17} & 67.1\B\\
					\specialrule{.15em}{0em}{0em}
				\end{tabular}
			\end{adjustbox}
			\caption{Quantitative results on the test set of CamVid. We report mean IoU (mIoU). Note that our method uses MobileNet-v2 as the encoder network.
				\label{table:camvid}}
		\end{center}
		%\vskip -0.3in
	\end{table}
	
\begin{figure*}[t]
		\centering
		\resizebox{1.\textwidth}{!}{\begin{tabular}{ccccc}
				\subfloat{\includegraphics[width = 0.19\linewidth]{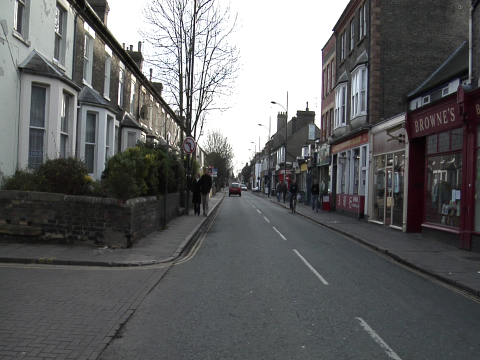}} &
				\subfloat{\includegraphics[width = 0.19\linewidth]{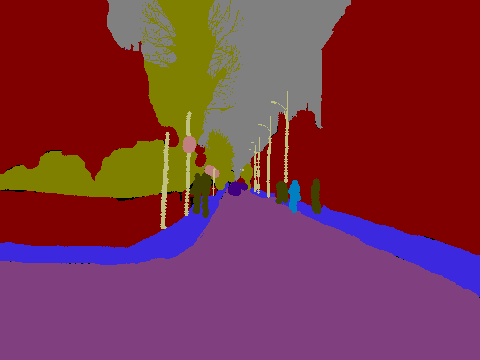}} &
				\subfloat{\includegraphics[width = 0.19\linewidth]{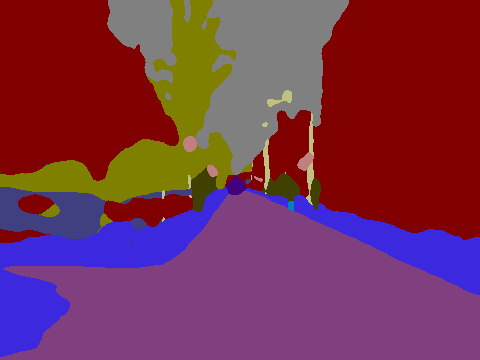}} &
				\subfloat{\includegraphics[width = 0.19\linewidth]{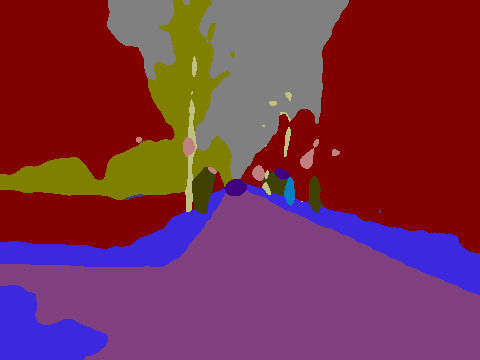}} &
				\subfloat{\includegraphics[width = 0.19\linewidth]{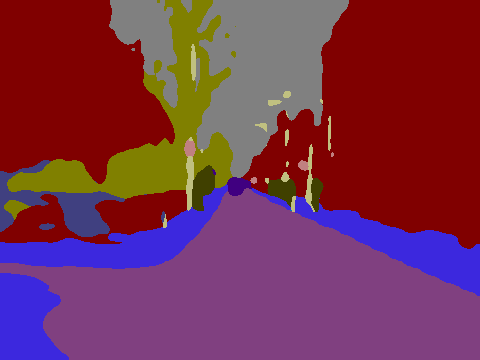}} \\[-0.15in]
				\subfloat{\includegraphics[width = 0.19\linewidth]{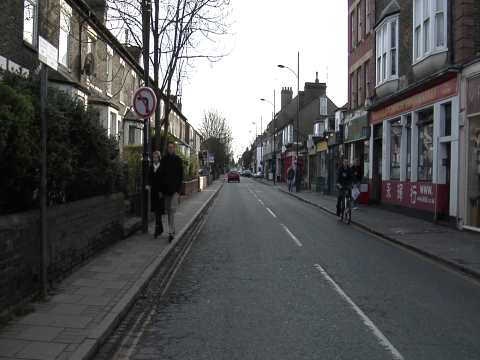}} &
				\subfloat{\includegraphics[width = 0.19\linewidth]{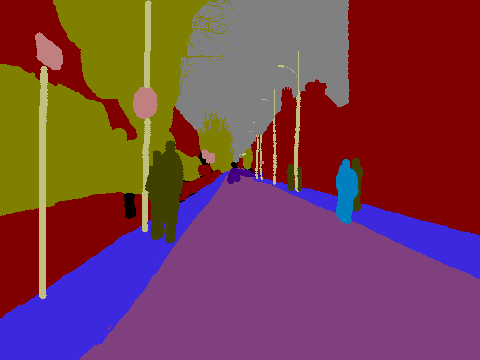}} &
				\subfloat{\includegraphics[width = 0.19\linewidth]{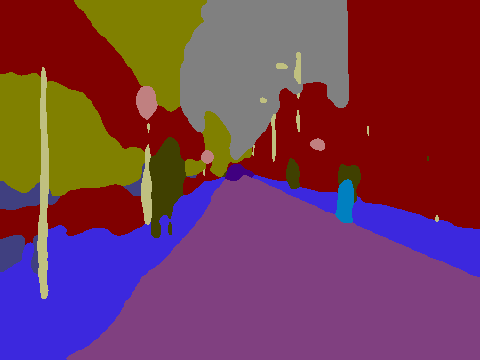}} &
				\subfloat{\includegraphics[width = 0.19\linewidth]{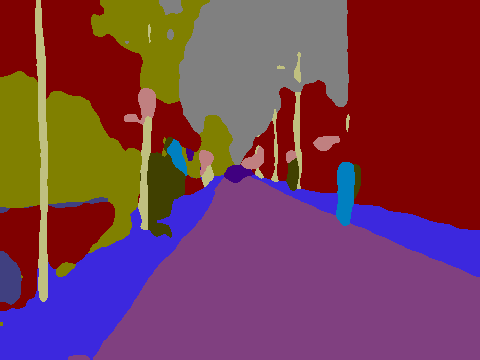}} &
				\subfloat{\includegraphics[width = 0.19\linewidth]{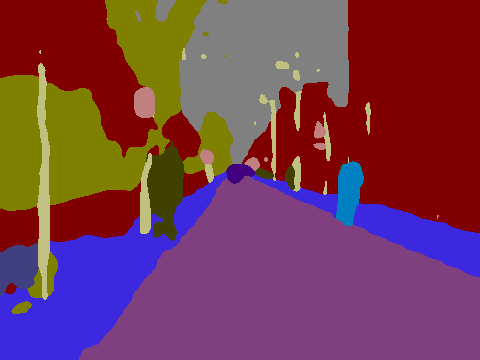}} \\[-0.15in]
				\subfloat{\includegraphics[width = 0.19\linewidth]{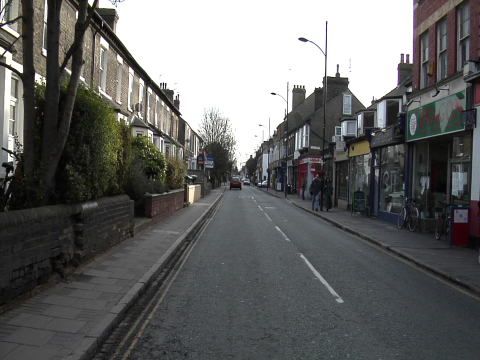}} &
				\subfloat{\includegraphics[width = 0.19\linewidth]{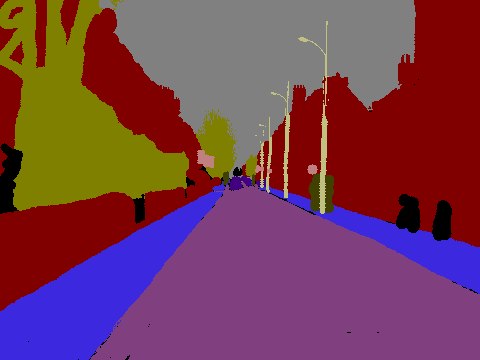}} &
				\subfloat{\includegraphics[width = 0.19\linewidth]{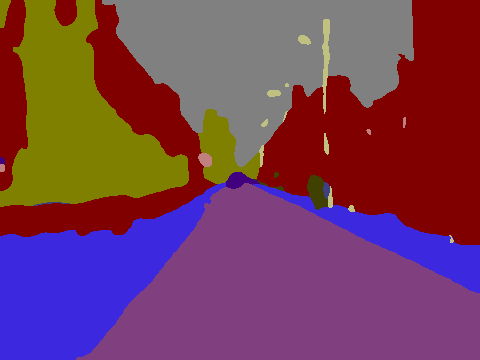}} &
				\subfloat{\includegraphics[width = 0.19\linewidth]{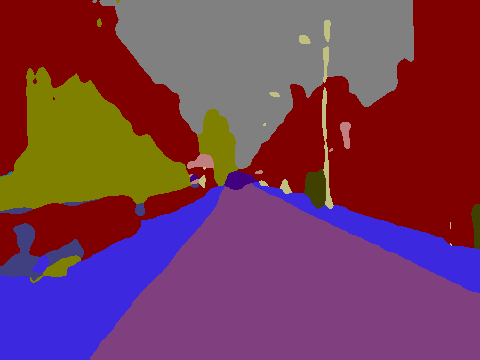}} &
				\subfloat{\includegraphics[width = 0.19\linewidth]{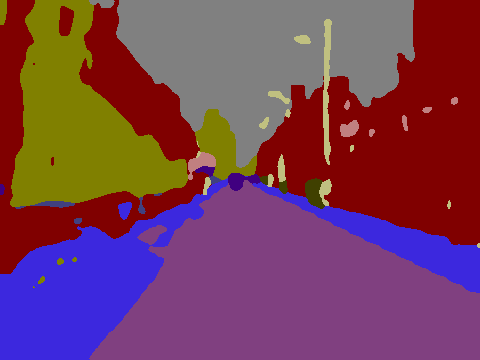}} \\[-0.15in]
				\subfloat{\includegraphics[width = 0.19\linewidth]{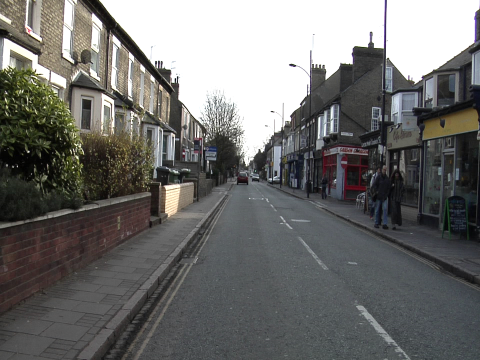}} &
				\subfloat{\includegraphics[width = 0.19\linewidth]{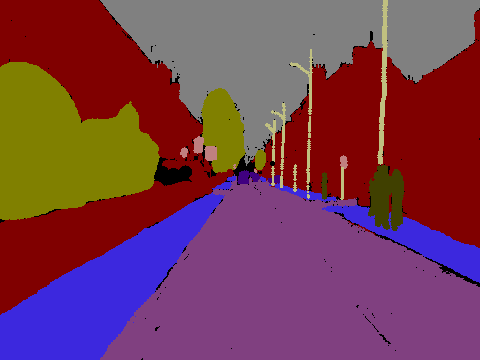}} &
				\subfloat{\includegraphics[width = 0.19\linewidth]{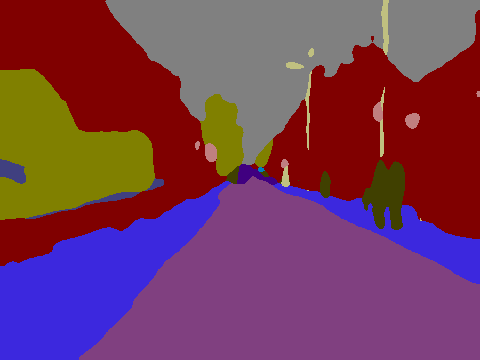}} &
				\subfloat{\includegraphics[width = 0.19\linewidth]{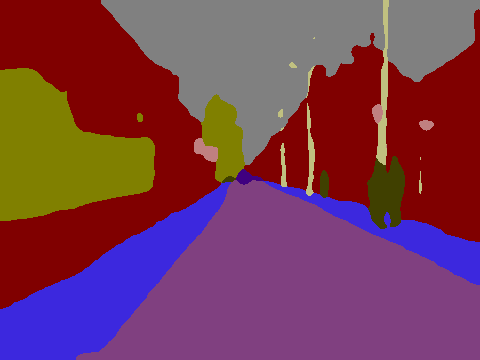}} &
				\subfloat{\includegraphics[width = 0.19\linewidth]{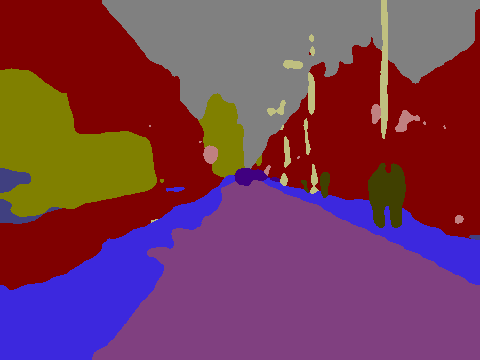}} \\[-0.15in]
				\subfloat{\includegraphics[width = 0.19\linewidth]{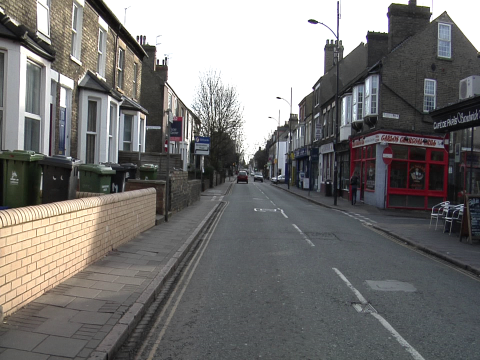}} &
				\subfloat{\includegraphics[width = 0.19\linewidth]{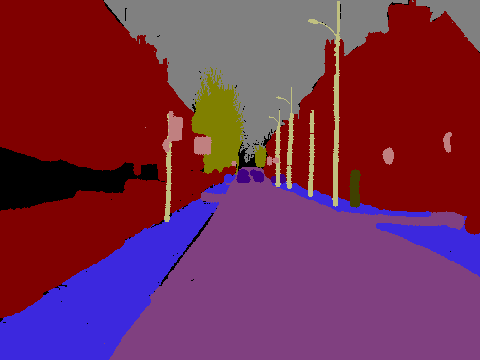}} &
				\subfloat{\includegraphics[width = 0.19\linewidth]{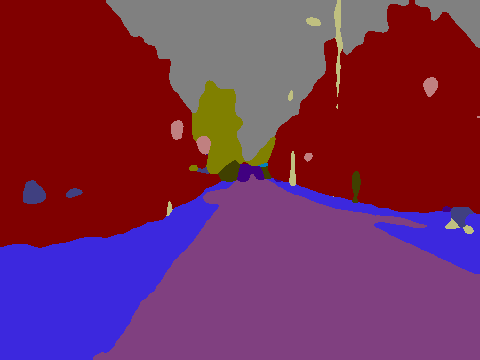}} &
				\subfloat{\includegraphics[width = 0.19\linewidth]{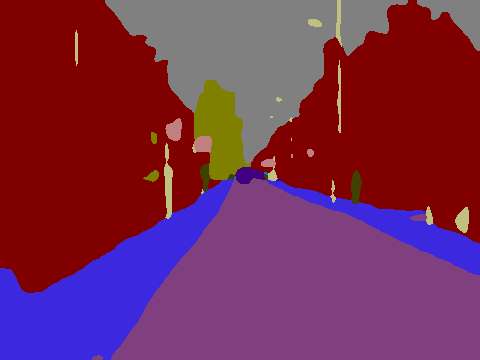}} &
				\subfloat{\includegraphics[width = 0.19\linewidth]{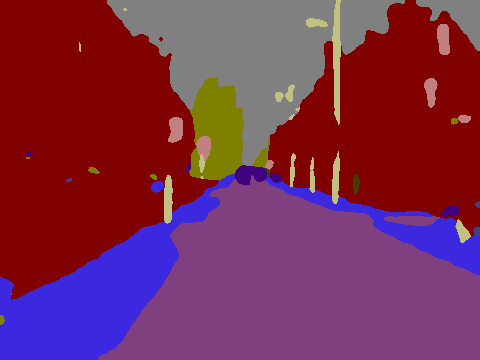}} \\
				Image & GT & Per-Frame & w/ cell0 & w/ cell1
		\end{tabular}}
		%\vskip -0.1in
		\caption{Inference results on the test set of CamVid.\label{fig:cv-res}}
		%\vskip -0.15in
	\end{figure*}

	We further visualise a few qualitative examples in Fig.~\ref{fig:cv-res}. The dynamic cell enables the network to effectively propagate information about thin structures, such as poles, which makes the resultant segmentation masks consistent in contrast to the per-frame baseline (rows $1{-}5$). Furthermore, the multi-frame segmentation network is able to track objects across neighbouring frames (rows $1{-}2$).

	\subsubsection*{CityScapes Results}
	We include the validation results of two discovered cells on CityScapes in Table~\ref{table:cs}. Once again, both dynamic cells are able to outperform the per-frame baseline by $1.2\%$. Furthermore, our models achieve favourable results in comparison to other video segmentation methods, all of which employ significantly larger backbones and, with the exclusion of Li~\etal~\cite{LiSL18}, all rely on the optical flow computation. Note also that Gadde~\etal~\cite{GaddeJG17} improved over their respective static baseline by $1.2\%$, too, while introducing a non-negligible overhead of $40$ms; and Li~\etal~\cite{LiSL18} compromised more than $3\%$ of the baseline score in order to reduce the latency. In contrast, we overcame our static baseline and decreased the runtime (Table~\ref{table:char}).  
	
\begin{figure*}[t]
		\centering
		\resizebox{1.\textwidth}{!}{\begin{tabular}{ccccc}
				\subfloat{\includegraphics[width = 0.19\linewidth]{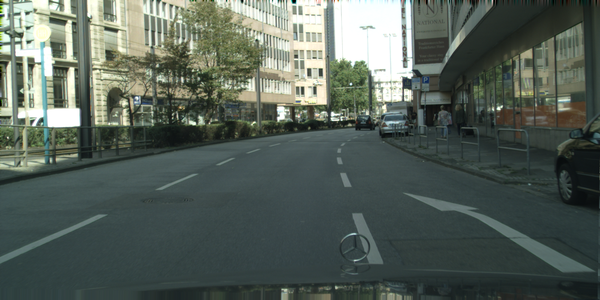}} &
				\subfloat{\includegraphics[width = 0.19\linewidth]{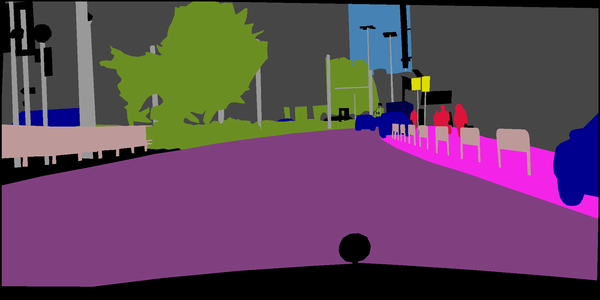}} &
				\subfloat{\includegraphics[width = 0.19\linewidth]{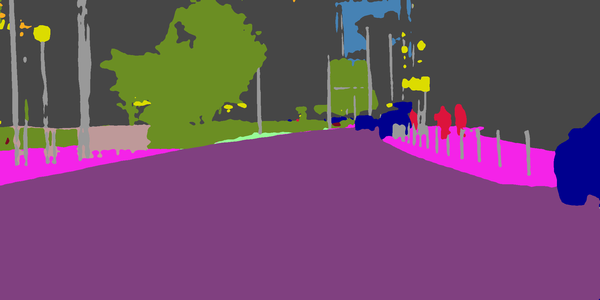}} &
				\subfloat{\includegraphics[width = 0.19\linewidth]{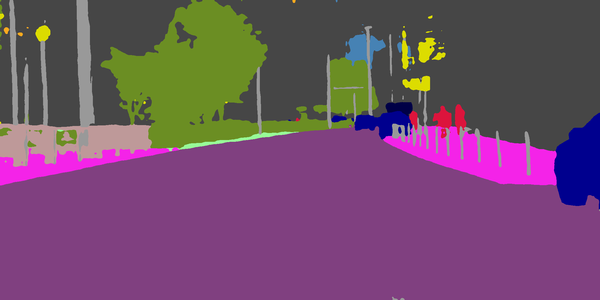}} &
				\subfloat{\includegraphics[width = 0.19\linewidth]{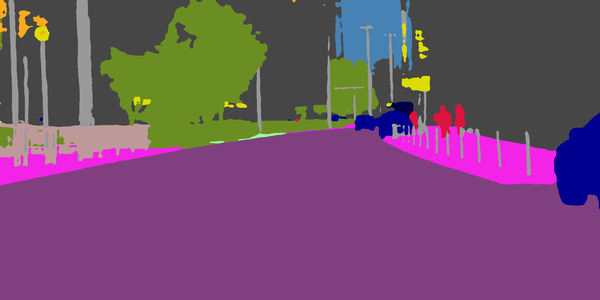}} \\[-0.15in]
				\subfloat{\includegraphics[width = 0.19\linewidth]{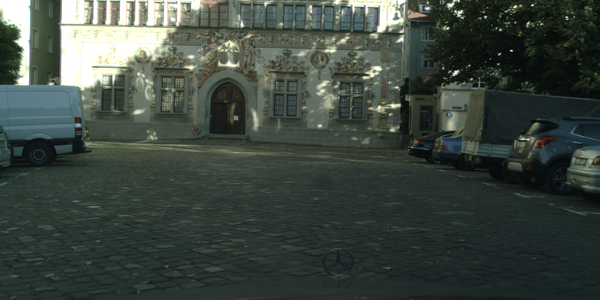}} &
				\subfloat{\includegraphics[width = 0.19\linewidth]{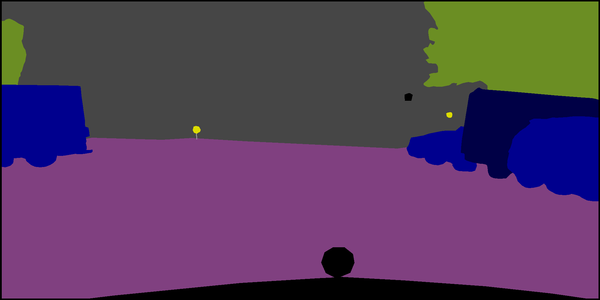}} &
				\subfloat{\includegraphics[width = 0.19\linewidth]{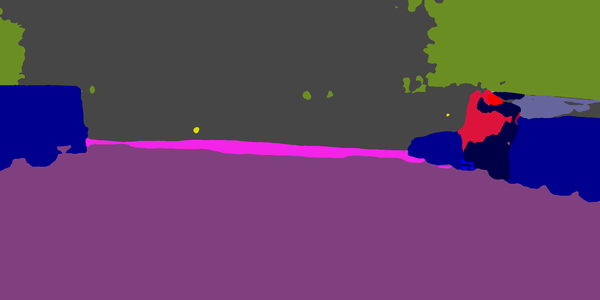}} &
				\subfloat{\includegraphics[width = 0.19\linewidth]{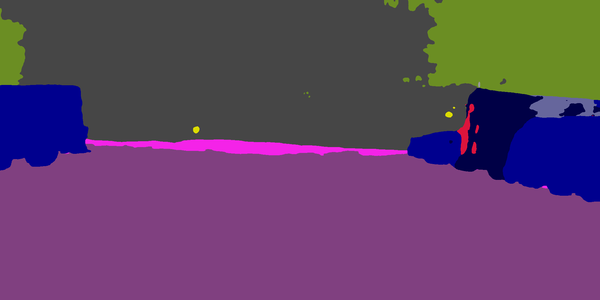}} &
				\subfloat{\includegraphics[width = 0.19\linewidth]{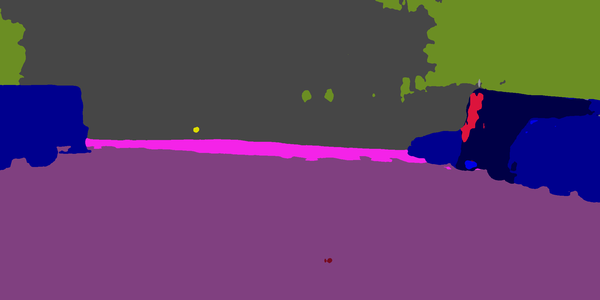}} \\[-0.15in]
				\subfloat{\includegraphics[width = 0.19\linewidth]{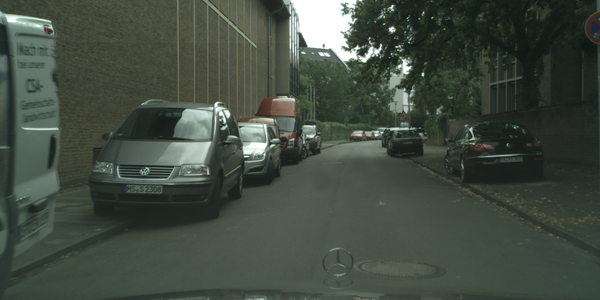}} &
				\subfloat{\includegraphics[width = 0.19\linewidth]{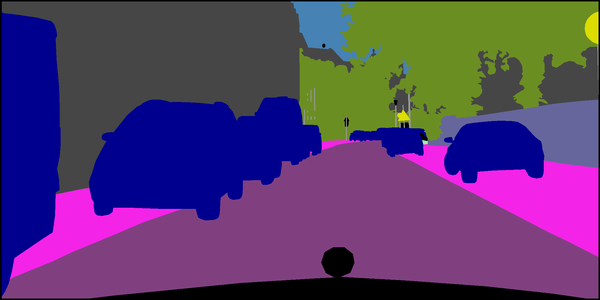}} &
				\subfloat{\includegraphics[width = 0.19\linewidth]{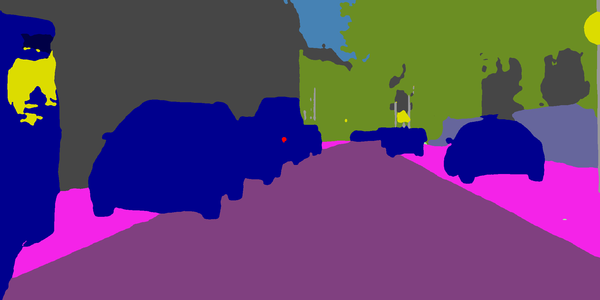}} &
				\subfloat{\includegraphics[width = 0.19\linewidth]{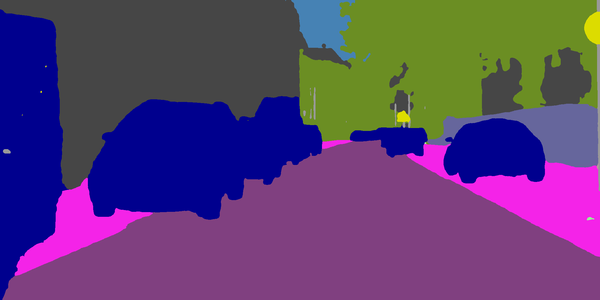}} &
				\subfloat{\includegraphics[width = 0.19\linewidth]{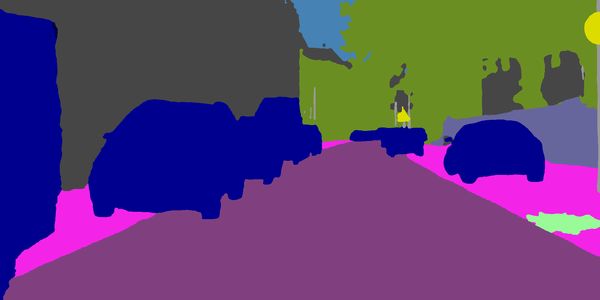}} \\[-0.15in]
				\subfloat{\includegraphics[width = 0.19\linewidth]{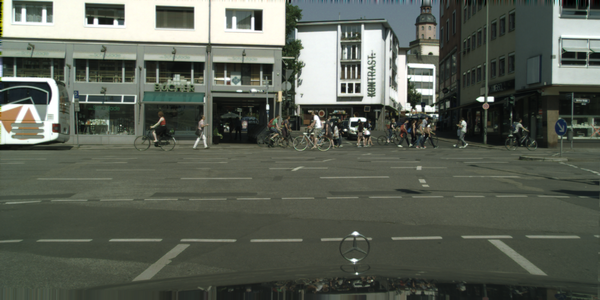}} &
				\subfloat{\includegraphics[width = 0.19\linewidth]{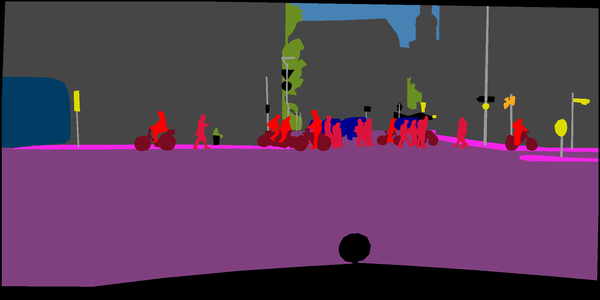}} &
				\subfloat{\includegraphics[width = 0.19\linewidth]{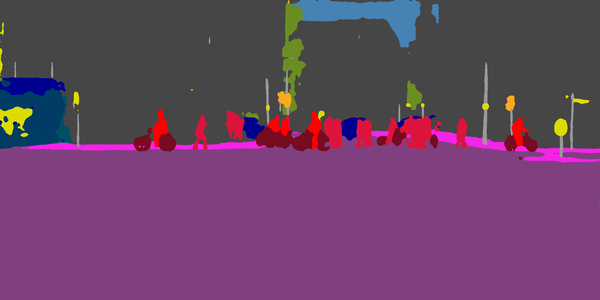}} &
				\subfloat{\includegraphics[width = 0.19\linewidth]{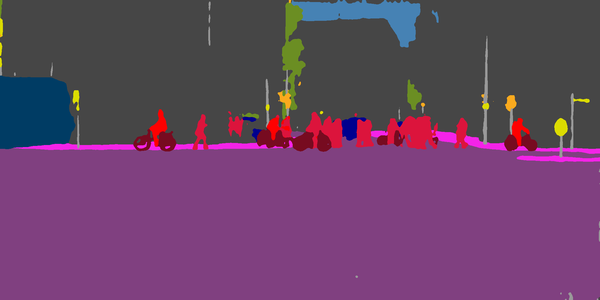}} &
				\subfloat{\includegraphics[width = 0.19\linewidth]{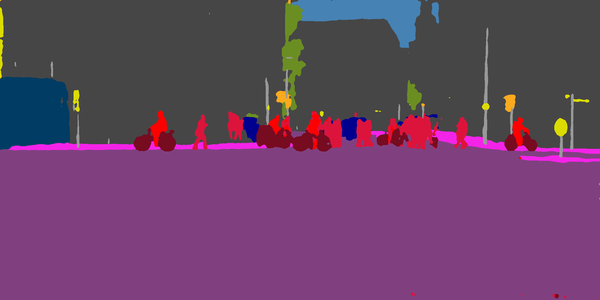}} \\[-0.15in]
				\subfloat{\includegraphics[width = 0.19\linewidth]{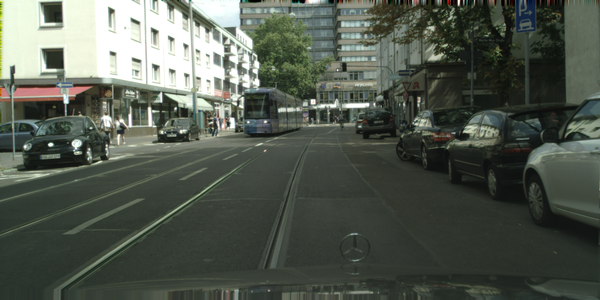}} &
				\subfloat{\includegraphics[width = 0.19\linewidth]{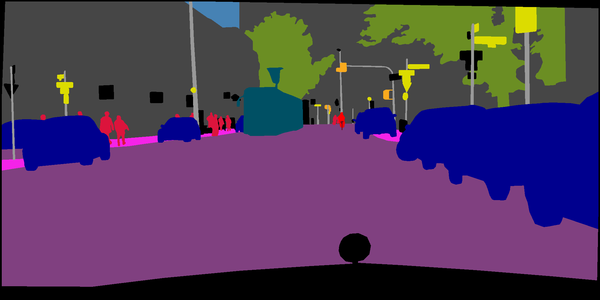}} &
				\subfloat{\includegraphics[width = 0.19\linewidth]{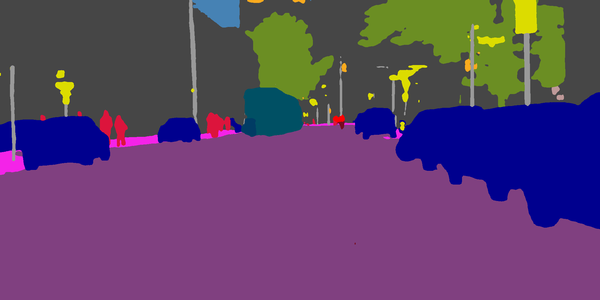}} &
				\subfloat{\includegraphics[width = 0.19\linewidth]{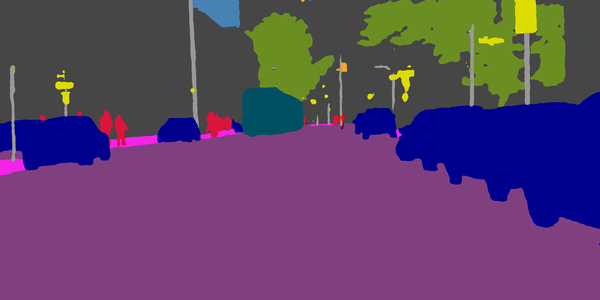}} &
				\subfloat{\includegraphics[width = 0.19\linewidth]{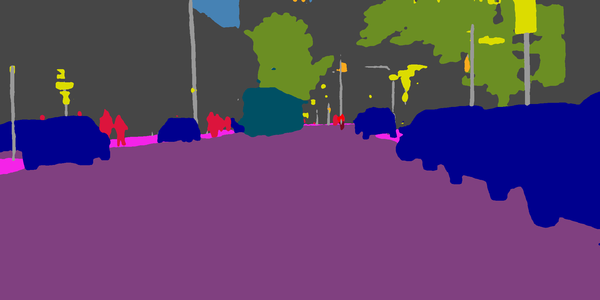}} \\
				Image & GT & Per-Frame & w/ cell2& w/ cell3
		\end{tabular}}
		%\vskip -0.1in
		\caption{Inference results on the validation set of CityScapes.\label{fig:cs-res}}
	\end{figure*}

	A few inference examples are visualised in Fig.~\ref{fig:cs-res}. As can be seen, the dynamic cells enhance the per-frame baseline results and identify partially occluded vehicles more accurately (rows $1{-}2$, $5$), while also avoiding misclassification of traffic signs at pixels with similar texture patterns (rows $2{-}4$).

	\begin{table}[htb]
	\begin{center}
		\begin{adjustbox}{max width=0.45\textwidth}
			\begin{tabular}{l|c}
				\specialrule{.15em}{0em}{0em}
				Method & mIoU,\% \T\B\\
				\specialrule{.1em}{0em}{0em}
				\hline
				per-frame baseline & 74.4 \T\B\\
				w/ cell2\tablefootnote{Test results: \url{https://bit.ly/2FrZ8jM}} & 75.6 \T\B\\
				w/ cell3\tablefootnote{Test results: \url{https://bit.ly/2HyoVcb}}& 75.6 \T\B\\
				\hline
				GRFP(5)~\cite{NilssonS18} & 69.5 \T\B\\
				Xu~\etal~\cite{XuFYL18} & 70.4 \T\B\\
				Li~\etal~\cite{LiSL18} & 76.8 \T\B\\
				Gadde~\etal~\cite{GaddeJG17} & 80.6 \B\\
				\specialrule{.15em}{0em}{0em}
			\end{tabular}
		\end{adjustbox}
		\caption{Comparison with other video segmentation approaches on the val set of CityScapes. Note that our method uses MobileNet-v2 as the encoder network.
			\label{table:cs}}
	\end{center}
	%\vskip -0.25in
\end{table}

	\subsection{Details of Discovered Architectures}
	
    We include characteristics of our networks together with numbers reported by others in Table~\ref{table:char}. As evident, our dynamic segmentation approach is superior to others in terms of its latency and compactness. Concretely, all our architectures contain at most $3.4$M parameters while having an average per-frame runtime of $50$ms on high-resolution $2048{\times}1024$ images. This is possible due to both the network design and the exclusion of the optical flow computation.

	\begin{table}[htb!]
		\begin{center}
			\begin{adjustbox}{max width=0.45\textwidth}
				\begin{tabular}{l|c|c|c|c}
					\specialrule{.15em}{0em}{0em}
					Method & GPU & Input Size & Param.,M & Avg. RT,ms \T\B\\
					\specialrule{.1em}{0em}{0em}
					\hline
				Baseline & 1080Ti & $2048{\times}1024$ & $\textbf{2.85}$ & 92.4$\pm$0.3 \T\B\\
				w/ cell0 & 1080Ti & $2048{\times}1024$ & 3.35 & 51.5$\pm$1.8\T\B\\
				w/ cell1 & 1080Ti & $2048{\times}1024$ & 3.19 & 52.6$\pm$1.8\T\B\\
				w/ cell2 & 1080Ti & $2048{\times}1024$ & 3.24 & 51.5$\pm$1.9\T\B\\
				w/ cell3 & 1080Ti & $2048{\times}1024$ & 3.30 & \textbf{50.5$\pm$1.9}\T\B\\
				\hline
				GRFP~\cite{NilssonS18} & TitanX & $512{\times}512$ & $>40$ & 685\T\B\\
				Li~\etal~\cite{LiSL18} & $-$ & $2048{\times}1024$ & $>40$ & 171 \T\B\\
			    Gadde~\etal~\cite{GaddeJG17} & TitanX & $2048{\times}1024$ & $>60$ & 3040\B\\
			    \specialrule{.15em}{0em}{0em}
				\end{tabular}
			\end{adjustbox}
			\caption{Number of parameters and average runtime (RT) comparison between different models. To compute average runtime with dynamic cells, we use the baseline on the first frame and the dynamic cell on the rest. Where possible, we report same characteristics for other methods. 
			\label{table:char}}
		\end{center}
		%\vskip -0.35in
	\end{table}

\begin{figure*}[t]
		\centering
		\resizebox{1.\textwidth}{!}{\begin{tabular}{cccc}
		    \multicolumn{4}{c}{\subfloat{\includegraphics[width=1.2\linewidth]{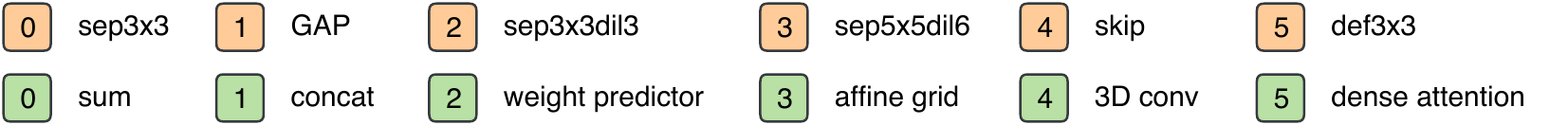}}}\\
		    \subfloat{\includegraphics[height=48mm]{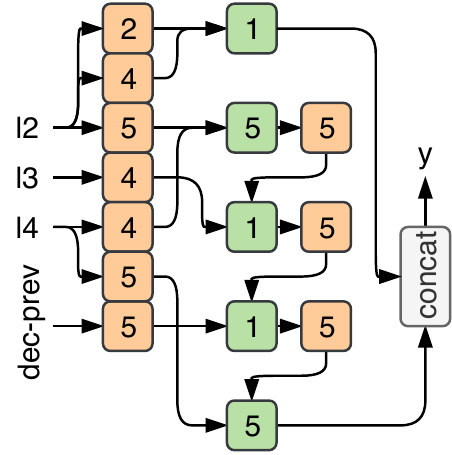}} &
			\subfloat{\includegraphics[height=48mm]{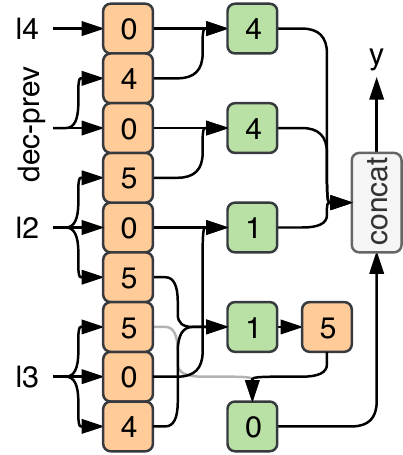}} &
			\subfloat{\includegraphics[height=48mm]{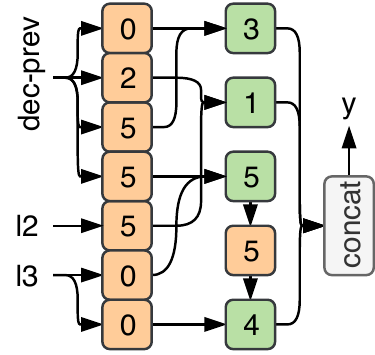}} &
			\subfloat{\includegraphics[height=48mm]{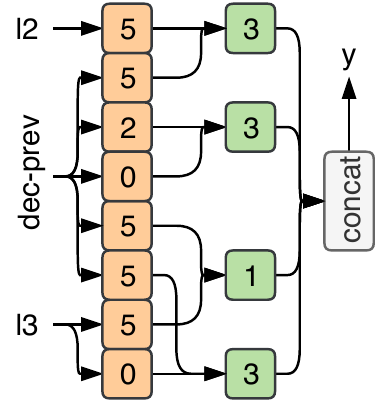}} \\
			cell0&cell1&cell2&cell3
		\end{tabular}}
		\caption{Visualisation of the discovered cells. Orange blocks represent operations and green blocks represent aggregation operations. Numbers inside blocks are operation identifiers as in Tables~\ref{table:ops} and~\ref{table:agg-ops}.}
		\label{fig:vis-archs}
	\end{figure*}

	All the trained cells are visualised in Fig.~\ref{fig:vis-archs}. Notably, layers with deformable convolution are present in all architectures. To propagate information from the previous frame, each cell exploits the $dec$ output instead of $layer~4$. All the cells prefer aggregating outputs via channel-wise concatenation with {\em cell0} also relying on dense attention, and {\em cell3} -- on affine transformation with bilinear sampling. In addition, {\em cell1} and {\em cell2} employ $3$D convolution in order to capture information between various inputs.
	
	\section{Discussion \& Conclusions}
	It is still an open question of what is the optimal way of propagating and extracting information across video frames. While a straightforward solution involving the optical flow allows to achieve solid results, it possesses several disadvantages that stem from the limitations of the optical flow itself and ultimately limit the ability of the network to adapt to novel frames. Furthermore, computations involving the optical flow cause a significant overhead, prohibiting the final system from being deployed in real-time.
	
	In this work, instead of manually enhancing static segmentation networks with dynamic components, we proposed an automatic approach based on neural architecture search methods. Such automation have multiple benefits as it explores a large pool of networks and finds best-performing ones on the given dataset. In a broader sense, starting from a static per-frame segmentation network, we showcased a way of generalising existing solutions without any reliance on the optical flow. In particular, we extended the static baseline with a dynamic cell, the design of which is automatically discovered with the help of reinforcement learning. The best discovered cells improve the baseline by more than $1\%$ at the same time leading to significant memory and latency savings. Concretely, two discovered cells on CityScapes reach $75.6\%$ mean IoU and require only $50$ms on average to process a $2048{\times}1024$ frame. While the proposed methodology relies on the static baseline, we expect that omitting that requirement and searching for a video segmentation network end-to-end would be an interesting problem to consider in the future work.
	
	\section*{Acknowledgements}
	VN, CS, IR's participation in this work were in part supported by ARC Centre of Excellence for Robotic Vision. CS was also supported by the GeoVision CRC Project.

	\section*{Appendix}
	
	\section{Training Details of Static Baseline}
	\label{static:train}
	The static baseline that we consider in the main text is {\em arch2} from~\cite{abs-1810-10804}, which we pre-train on CamVid~\cite{BrostowFC09} and CityScapes~\cite{CordtsORREBFRS16}.
	
	We utilise the `poly' learning schedule~\cite{ChenPKMY18} with the initial learning rates of $5{e-}2$ and $1{e-}2$ for the encoder and the decoder, respectively. As in~\cite{abs-1810-10804}, we set the weight for auxiliary losses to $0.3$.
	
	On CityScapes, we train for $1000$ epochs with mini-batches of $28$ examples each randomly scaled with the scale factor in range of $[0.5, 2.0]$ and randomly cropped to $800{\times}800$ with each side zero-padded accordingly. On CamVid, we train for $2000$ epochs with mini-batches of $32$ examples each randomly scaled with the scale factor in range of $[0.5, 2.0]$ and randomly cropped to $600{\times}600$ with each side zero-padded accordingly.
	
	{\small
		\bibliographystyle{ieee}
		\bibliography{egbib}
	}
	
\end{document}